\documentclass[10pt,twocolumn,letterpaper]{article}

\PassOptionsToPackage{usenames,dvipsnames}{xcolor}

\usepackage{style/iccv}
\usepackage{times}
\usepackage{epsfig}
\usepackage{graphicx}
\usepackage{amsmath}
\usepackage{amssymb}

\usepackage[dvipsnames]{xcolor}
\usepackage{xspace}

\usepackage{caption}
\usepackage{subcaption}
\usepackage{svg}

\usepackage{soul}

\usepackage{booktabs}
\usepackage[normalem]{ulem}
\useunder{\uline}{\ul}{}
\usepackage[most]{tcolorbox}
\tcbset{on line, 
        boxsep=3pt, left=0pt,right=0pt,top=0pt,bottom=0pt,
        colframe=white,colback=lightgray,  
        }

\usepackage[pagebackref=true,breaklinks=true,colorlinks,bookmarks=false]{hyperref}

\iccvfinalcopy

\usepackage{circledsteps}

\renewcommand{\arraystretch}{1.25}

\newcommand{\dcpos}[1]{\textcolor[HTML]{73CF9B}{#1}}
\newcommand{\dcneg}[1]{\textcolor[HTML]{E37C74}{#1}}
\newcommand{\dcneu}[1]{\textcolor[HTML]{646464}{#1}}

\newcommand{\xhdr}[1]{\vspace{4pt}\noindent\textbf{#1}}
\newcommand{\xhdrflat}[1]{\noindent\textbf{#1}}
\newcommand{\uhdr}[1]{\vspace{4pt}\noindent\underline{\smash{#1}}}

\newcommand{\myquote}[1]{``\emph{#1}''}

\newcommand{\tabref}[1]{Tab.~\ref{#1}\xspace}
\newcommand{\figref}[1]{Fig.~\ref{#1}\xspace}
\newcommand{\secref}[1]{Sec.~\ref{#1}\xspace}
\renewcommand{\eqref}[1]{Eq.~\ref{#1}\xspace}

\renewcommand{\etal}{\emph{et al.}\xspace}
\renewcommand{\eg}{\emph{e.g.}\xspace}
\renewcommand{\ie}{\emph{i.e.}\xspace}

\newcommand{\vlslice}{\texttt{VL}\textsl{Slice}\xspace}
\definecolor{mygray}{RGB}{85,85,85}
\newcommand{\mycirc}[1]{\Circled[fill color=mygray, inner color=white, outer color=mygray]{#1}}

\begin{document}

\title{\vlslice: Interactive Vision-and-Language Slice Discovery}

\author{Eric Slyman\\
Oregon State University\\
Corvallis, OR, USA\\
{\tt\small slymane@oregonstate.edu}
\and
Minsuk Kahng\\
Google Research\\
Atlanta, GA, USA\\
{\tt\small kahng@google.com}
\and
Stefan Lee\\
Oregon State University\\
Corvallis, OR, USA\\
{\tt\small leestef@oregonstate.edu}
}
\maketitle

\begin{abstract}
Recent work in vision-and-language demonstrates that large-scale pretraining can learn generalizable models that are efficiently transferable to downstream tasks.
While this may improve dataset-scale aggregate metrics, analyzing performance around hand-crafted subgroups targeting specific bias dimensions reveals systemic undesirable behaviors. 
However, this subgroup analysis is frequently stalled by annotation efforts, which require extensive time and resources to collect the necessary data.
Prior art attempts to automatically discover subgroups to circumvent these constraints but typically leverages model behavior on existing task-specific annotations and rapidly degrades on more complex inputs beyond ``tabular'' data, none of which study vision-and-language models.
This paper presents \vlslice, an interactive system enabling user-guided discovery of coherent representation-level subgroups with consistent visiolinguistic behavior, denoted as vision-and-language slices, from unlabeled image sets. 
We show that \vlslice enables users to quickly generate diverse high-coherency slices in a user study (n=22) and release the tool publicly\footnote{\href{https://github.com/slymane/vlslice}{\href{https://github.com/slymane/vlslice}{https://github.com/slymane/vlslice}}}.
\vspace{-6.5pt}
\end{abstract}

\section{Introduction}
\label{sec:intro}

Large-scale vision-and-language models trained on curated \cite{multi-vilbert,uniter,lxmert} and web-scrapped \cite{clip,align,pali} data have led to significant improvements over task-specific models when transferred to downstream tasks in terms of aggregate metrics. However, researchers probing these models on hand-curated datasets have revealed problematic behaviors and well-known biases \cite{ross_measuring_2020,srinivasan_worst_2021} learned during pretraining -- \eg biases with respect to perceived gender, skin tone,\footnote{We use the term `skin tone' rather than `race' as race is a socially constructed identity that can span a range of phenotypic features.} and occupation. These biases can lead to disparate representational performance for population subgroups, resulting in poor prediction quality for downstream applications such as image captioning \cite{hendricks2016women,wang_measuring_2022} and search \cite{algorithmsofopression}.

In the standard paradigm for bias analysis in vision-and-language models, researchers query and analyze a set of images that potentially exhibit bias. They often select a subject population of interest, some specific subgroups of those subjects to analyze, and a bias dimension to measure against the model \cite{ross_measuring_2020,srinivasan_worst_2021}. For example, Ross \etal~\cite{ross_measuring_2020} choose images of people as their subject population, label these based on perceived gender and skin tone categories, and measure model-predicted affinities between the labeled image subsets and text describing occupations or (un)pleasantness.

To effectively support analysis of such image sets for researchers, the set of images returned for the subject population subsets should be \textbf{large}, \textbf{coherent}, and \textbf{representative} -- \ie containing enough images to make statistically significant statements, capturing a well-defined visual concept, and covering the full diversity of visual presentation for the selected concept rather than an arbitrary subset. Without these, the biases may simply be noise (\st{large}), be obscured by effects from images outside the intended subject group being included (\st{coherent}), or be the result of some intersectional bias captured in the subset that is not consistent across the whole expression of the visual concept (\st{representative}).

Collecting and labeling appropriate image sets that fulfill these properties can be an arduous task. Despite this, manual annotation of static datasets along predefined subgroup and bias dimensions is the standard practice \cite{karkkainenfairface,holstein2019improving}. This data collection methodology is expensive to perform -- effectively limiting broad bias-auditing to high-resource institutions. Further, the one-off nature of this labeling process limits the scope of testing to pre-identified biases and does not account for how concepts may shift in visual expression or cultural convention over time.

Several methods have been proposed to automatically discover biased \myquote{slices} of data which share similar input attributes and exhibit consistent responses from machine learning models \cite{slicefinder,sliceline,domino,spotlight,sohoni_subclass_2022,multiaccuracy}. These Slice Discovery Methods (SDMs) have typically been deployed in tabular input settings where individual input dimensions are semantically meaningful. While some recent work has explored extending SDMs to more complex inputs like images \cite{domino,spotlight,sohoni_subclass_2022,multiaccuracy}, these require task-specific annotations to evaluate the model -- making them unsuitable to auditing general vision-and-language alignment models.

\begin{figure}[t]
     \centering
     \includegraphics[trim=1.95in 1.50in 1.95in 1.45in,clip,width=\linewidth]{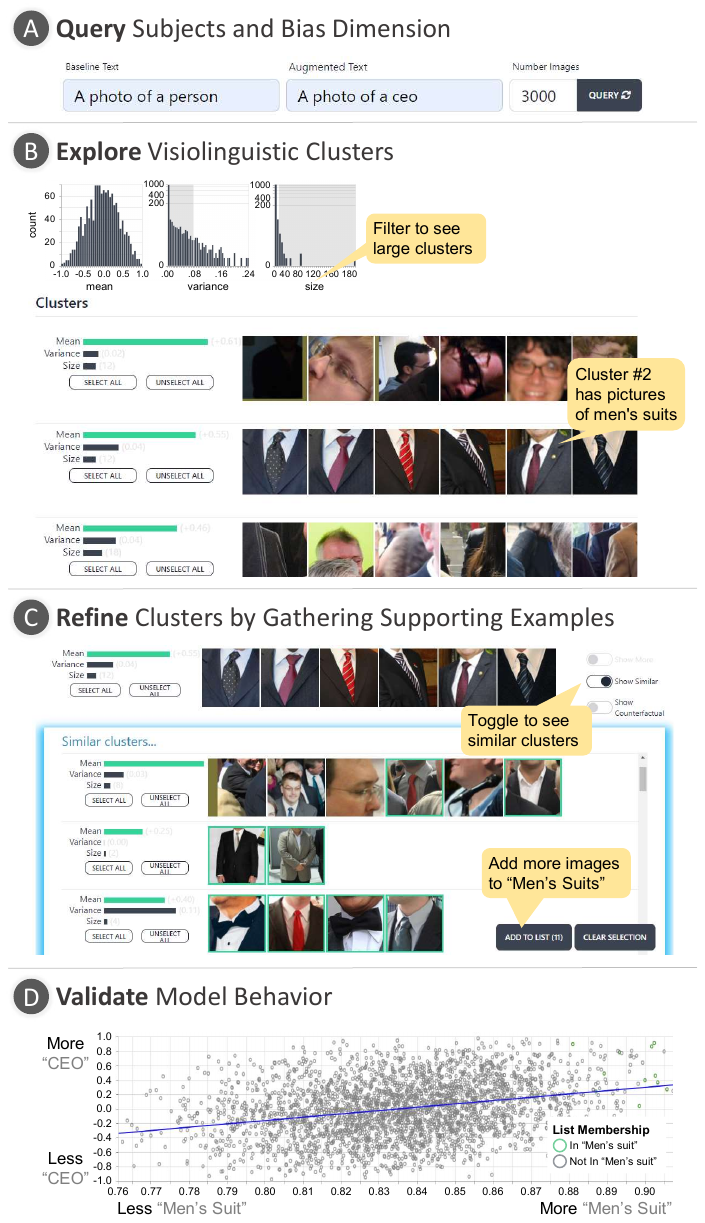}
     \caption{An example user workflow with \vlslice. The user workflow begins with writing a \textbf{(A) Query} to the model then \textbf{(B) Explor}ing the resulting visiolinguistic clusters to find interesting candidates to begin building a slice from. Once users identify a hypothesis, they can  \textbf{(C) Refine} the clusters by gathering additional samples in a human-in-the-loop manner with \vlslice recommending similar and counterfactual examples to add to the clusters. Finally, users can \textbf{(D) Validate} the bias behavior of the model on this slice.}
     \label{fig:vlslice}
\end{figure}

To improve this workflow, we propose \vlslice, an interactive system to discover vision-and-language slices from unlabeled collections of images. \vlslice consists of four primary stages of user-driven interaction with a vision-and-language alignment model of interest as depicted in the system overview in \figref{fig:vlslice}.
\mycirc{A} First, users write a query defining a subject population of interest (\eg, \myquote{person}, \myquote{car}) and bias dimension to measure (\eg, \myquote{intelligent}, \myquote{fast}) which is submitted to \vlslice to select from a large set of unlabeled images down to a subset of subject-relevant images, then cluster those images by visual similarity and alignment with the bias dimension.
\mycirc{B} Second, users are displayed the clusters generated by \vlslice and can search, filter, and sort those clusters in (un)directed searches to identify and capture candidate slices (\eg, \myquote{people wearing suits}, \myquote{red cars}).
\mycirc{C} Next, users interact with \textit{VLSlice} in a loop viewing recommended similar and counterfactual clusters to their slice to gather more coherent and representative samples.
\mycirc{D} Finally, users can view a plot that shows the relationship between the slice they formed and the bias term across the entire subject population of interest, validating if biased model behavior is demonstrated in the slice.

We demonstrate that \vlslice enables users to quickly generate diverse high-coherency slices in a between-subjects user study ($n{=}22$), contrasting with a control interface mimicking a linear unguided image search. From this study, we present both qualitative and quantitative support, and discuss emergent user interaction paradigms. We choose to study CLIP \cite{clip} as a representative model of contemporary methods in large-scale pretraining and self-supervision for image-text alignment.

\section{Related Work}
\label{sec:relwork}

\xhdrflat{Vision and Language Bias.}
Both vision and language models are independently known to harbor biases leading to representational harm. For example, gender and skin tone \cite{gendershades} in vision systems, associating racial minorities with animals \cite{nytgoogle}, gendering professions in language models \cite{bolukbasi2016ManIT}, and a litany of others documented in \cite{languagebias,mitigatinggenderbias}. Multimodal vision-and-language models are not exempt from these tendencies \cite{nytfacebook,hendricks2016women,algorithmsofopression} and may in fact exacerbate them \cite{srinivasan_worst_2021}.

Many modern, high-performing vision-and-language alignment models are pretrained on scraped internet data \cite{clip,align,pali} -- a strategy that improves performance but has been shown to teach models \myquote{misogyny, pornography, and malignant stereotypes} \cite{birhane_multimodal_2021} across multiple studies \cite{birhane_multimodal_2021,ross_measuring_2020,srinivasan_worst_2021}. As discussed in \secref{sec:intro}, studying these biases often requires labelling thousands of images for specific preconceived subject groups and bias dimensions (\eg gender \cite{srinivasan_worst_2021} and skin tone \cite{ross_measuring_2020,hirota2022quantifying}, or emotion \cite{mohamed2022itisokay}). Our proposed interactive method allows for more open-ended exploration of bias and reduces the burden of collecting relevant subject group image sets.

\newpage
\xhdr{Slice Discovery \& Exploration.}
Slice discovery methods (SDMs) attempt to find critical subgroups (or \textit{slices}) of data with common input properties and consistent predictions (often, mis-predictions) with respect to some model. Much of this work is developed for tabular data settings where slices are defined based on categorical (e.g., gender, occupation)~\cite{slicefinder,sliceline} and numerical attributes (e.g., age, duration in days)~\cite{cabrera_fairvis_2019}. In these settings, identifying semantically coherent subsets is more straightforward than in high-dimensional perceptual data like images where individual input dimensions are non-semantic. Once slices are discovered, they can be interactively inspected to understand model behavior. For the case of slices that contain many mis-predicted examples, they can be leveraged to improve model performance, such as by augmenting the training set with additional examples within the slice~\cite{chen_slice-based_2019,slicetuner}.

Recent SDM work explores \myquote{unstructred} data like images, where each data item is not necessarily associated with structured attributes~\cite{domino,spotlight,sohoni_subclass_2022,multiaccuracy}. Commonly, these approaches attempt to extract semantically meaningful clusters from their high-dimensional embedding representations. For instance, Spotlight~\cite{spotlight} identifies contiguous representation space regions that contain data items with high loss. 
Domino~\cite{domino} additionally leverages a joint input-and-language embedding space to generate natural language descriptions of the extracted slices after input-space clustering, which can potentially aid analysis. \vlslice, in contrast, examines the relationship between two modalities rather than the nature of an input and a model's task specific predictions. We note that the visiolinguistic relationship captured by image-text alignment may be viewed as an image classification task with an extremely large and sparse label space (all natural language strings). Under this setting, \vlslice examines a sparsely labeled many-to-many relationship between inputs where an image (caption) may match many captions (images) but only a single relationship is known. No SDMs exist that examine relationships of this noisy multimodal nature.

Moreover, automated SDMs for unstructured data have critical limitations. Because of the nature of unsupervised methods, extracted clusters cannot perfectly align with semantically meaningful concepts, bias, and human knowledge~\cite{lee2012ivisclustering}. Therefore, results from automated methods need to be inspected and refined by human users (e.g., merge, split, add) to identify slices that capture the necessary concepts while meeting the desired properties such as coherency and representativeness. \vlslice mitigates the noise in multimodal relationships and improves alignment to semantically meaningful concepts by providing tooling for human-in-the-loop slice discovery and refinement.

The field of \textit{visual analytics} has developed methods and tools that leverage the power of humans in data analysis~\cite{keim2008visual}. Visual analytics tools frequently serve to help users explore noisy slices returned by SDMs. FairVis~\cite{cabrera_fairvis_2019} allows users to discover subgroups that exhibit bias by interactively analyzing the clustering of tabular data. Zhao \etal~\cite{zhao_human---loop_2021} enables users to train a binary classifier for each slice through active learning, based on their analysis of the initial clustering of image patches. While most existing work targets a single modality,
Cabrera \etal~\cite{cabrera2022did} target image captioning tasks. In contrast to their focus on human's comprehensive sensemaking of non-grouped image sets, \vlslice aims to use human inputs minimally by letting human users start their analysis from clustered results.

\section{\vlslice}
We propose \vlslice, an interactive system to discover and build slices from large-scale unlabeled image sets with respect to a vision-and-language (ViL) model of interest. Specifically, our methodology is designed to test ViL models which produce image-text alignment scores and support clustering in the image feature space -- a common feature of many popular models (\eg CLIP \cite{clip}). 

The following subsections describe the technical details of \vlslice and provide a running example of how they support the user workflow depicted in \figref{fig:vlslice}. The workflow is roughly split into four parts -- \mycirc{A} query specification (\secref{sec:query}), \mycirc{B} exploration of presented clusters (\secref{sec:explore}), \mycirc{C} iterative slice refinement (\secref{sec:gather}), and \mycirc{D} validation of the observed phenomenon (\secref{sec:validate}). 

\subsection{Caption-based Querying} \label{sec:query}
Users must first write a query defining the domain of subjects (\eg, people, cars, houses) and bias dimension (\eg, CEO-like, fast, pleasant) they are interested in evaluating by specifying the baseline and augmented captions in the \vlslice interface, denoted respectively as $C_b$ and $C_a$. The interface will then define a \textit{working set} $\mathcal{I}_w$ of relevant images by filtering a large unlabelled image set to just the $k$ images most aligned with the baseline caption.  The choice of $k$ is left to the user and is a trade-off between precision and recall of subjects captured within the working set. We discuss this trade-off further in \secref{sec:limitations}.

\uhdr{Running Example} (\figref{fig:vlslice} \mycirc{A}). The user enters \myquote{A photo of a person} as the baseline caption $C_b$ with $k{=}3000$ to restrict the working set to 3000 \myquote{people}-images and \myquote{A photo of a \underline{CEO}} as the augmented caption $C_a$ to define a \myquote{CEO}-ness bias dimension to explore.

\xhdr{Measuring Affinity with the Augmented Caption.} Without loss of generality, we can consider vision-and-language affinity models to be functions $f(I,C)$ that compute some score reflecting if caption $C$ describes the contents of the image $I$. A natural approach for measuring the model's predicted affinity between each working set image $I_i\in\mathcal{I}_w$ and the bias dimension is then to compute the \emph{augmented caption similarity} $S^a_i{=}f(I_i, C_a)$. However, our initial experiments suggest this is not always sufficient.

\begin{figure}[t]
    \centering
    \includegraphics[trim=2.1in 4.1in 2.0in 4.0in,clip,width=\linewidth]{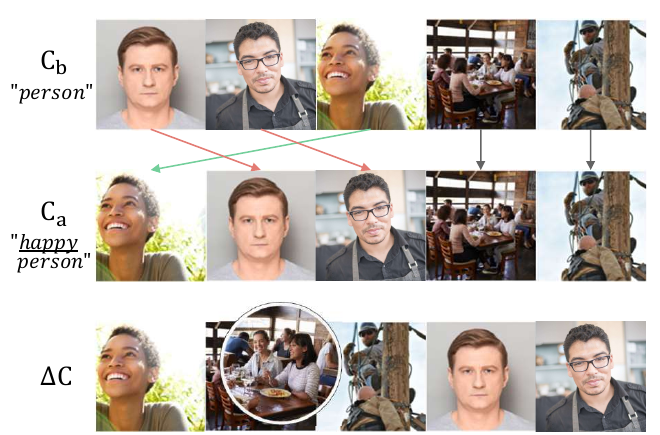}
    \vspace{-11pt}
    \caption{Sample rankings by baseline caption ($C_b$ = \myquote{person}), augmented caption ($C_a$ = \myquote{\underline{happy} person}), and $\Delta C$, with highest on the left. The change in percentile from $C_b$ to $C_a$ is shown with \dcpos{green arrows} for positive changes, \dcneg{red arrows} for negative, and \dcneu{gray arrows} for neutral. We enlarge the photo of people with smiling faces eating a meal. The rank of this photo does not change from $C_b$ to $C_a$ (4th), but increases (2nd) under $\Delta C$. Captions are prepended with \linebreak
    \myquote{A photo of a \rule{0.75cm}{0.15mm}} in practice.}
    \label{fig:deltac}
\end{figure}

In \figref{fig:deltac}, we consider cases where the augmented caption \emph{extends} the baseline caption, \eg \myquote{A photo of a person} vs.~\myquote{A photo of a {\underline{{happy}}} person}. In these cases, well-framed canonical images of the subject may retain higher scores under the augmented caption despite actually \emph{reducing} in magnitude compared to their affinity to the baseline caption -- due solely to a strong alignment with the subject. In this instance, we would like to disentangle the model-of-interest's learned notions of \myquote{person} from that of \myquote{happy} which we wish to examine.

To ameliorate this effect, we consider a measure based on the change in similarity from the baseline to augmented caption.
Analogous to the augmented caption similarity $S^a_i$, let $S^b_i{=}f(I_i, C_b)$ be the baseline caption similarity for image $I_i$. With $S^a$ and $S^b$ denoting the empirical distributions of these similarities over the working set $\mathcal{I}_w$, let $P^a_i{=}P(S^a \leq S^a_i)$ and $P^b_i{=}P(S^b \leq S^b_i)$ be the empirical percentile rank of caption similarity for image $I_i$ for the augmented and baseline captions respectively.

We can then define the \emph{change in augmented caption percentile} $\Delta C$ for image $I_i$ as%
\begin{equation}
     \Delta C(I_i) = P^a_i-P^b_i.
\end{equation}
Intuitively, images that report \emph{greater} affinity with the augmented caption than the baseline caption relative to other working set images will achieve a higher $\Delta C$. As shown in \figref{fig:deltac}, this results in reduced effect of canonical images. Further, this score is bounded between $\pm1$ and avoids forcing users to reason about absolute changes in affinity magnitude which may vary greatly in scale for different settings.

\subsection{Exploring Visiolinguistic Clusters} \label{sec:explore}

To assist users in identifying groups of related images with consistent affinity to the augmented caption, we display the working set images as clusters rather than a long list of individual images. These accelerate users' ability to find examples, prime them for identifying shared visual features, and provides a convenient bootstrap for forming slices. 

We form clusters using standard agglomerative clustering with average linkage stopping at distance threshold $dt=0.2$. To capture both visual similarity and bias-effect consistency, we define the linkage affinity between two images $I_i$ and $I_j$ as a combination of their visual dissimilarity $D_{\mbox{img}}$ and the difference between their $\Delta C$'s. Given a visual encoder $\Phi: I \rightarrow \mathbb{R}^d$ from our model-of-interest, we compute the visual cosine distance as
\begin{equation}
    D_{img}(I_i, I_j) = 1 - \Phi(I_i)^T\Phi(I_j) / \lVert\Phi(I_i)\rVert \lVert\Phi(I_j)\rVert, 
\end{equation}
the $\Delta C$-based affinity consistency distance as
\begin{equation}
    D_{\Delta C}(I_i, I_j) = \vert\Delta C(I_i) - \Delta C(I_j)\vert,
\end{equation}
and the overall linkage affinity between $I_i$ and $I_j$ as
\begin{equation}
    D(I_i, I_j) = a * D_{img}(I_i, I_j) + (1 - a) * D_{\Delta C}(I_i, I_j) 
\end{equation}
where $a$ controls the trade-off between clustering by visual similarity and affinity consistency. We set $a=0.95$.

By default, clusters are displayed in descending order by mean $\Delta C$. Each cluster is displayed with a set of sample images along with numeric attributes for cluster size and mean / variance of $\Delta C$ within the cluster. Additionally, histograms for these numeric attributes are displayed and users may filter the displayed clusters by specifying ranges of attribute values. Users may change the ordering of clusters by ranking on different attributes or by making a directed search over clusters with any arbitrary text (\eg, \myquote{glasses}). In this case, clusters are re-ranked by average image-text similarity from the model-of-interest. To construct slices, users create lists of images according to the properties enumerated in \secref{sec:intro}. Users may select individual image instances or entire clusters, then add-to an existing or new slice. Upon slice creation, users are prompted to provide a name according to the captured visual feature(s) and may change that name later as the slice is refined.

\uhdr{Running Example} (\figref{fig:vlslice} \mycirc{B}). The user interacts with the histogram filters to find large clusters with low variance in $\Delta C$ to drill-down on high-quality clusters likely to prompt some bias hypothesis. Then, they browse the sorted clusters to discover a visual concept corresponding to men's suits.

\subsection{User-guided Cluster Refinement} \label{sec:gather}

Once users have identified a slice they would like to explore by capturing some example images, our interface provides tooling to assist with discovering additional examples to improve the size of the sample and representation of the visual concept. Specifically, users may request \emph{similar} and \emph{counterfactual} clusters for a slice. We measure cluster similarity based on the cosine between cluster centroids. When \emph{similar} clusters are requested for a slice, we simply rank clusters by visual similarity to the slice and display the nearest 50 clusters. For \emph{counterfactual} clusters, we first filter to clusters which have mean $\Delta C$'s with opposing sign as the slice and then sort by visual similarity. 

\begin{figure}[t]
    \centering
    \includegraphics[trim=3.0in 4.1in 3in 4.1in,clip,width=\linewidth]{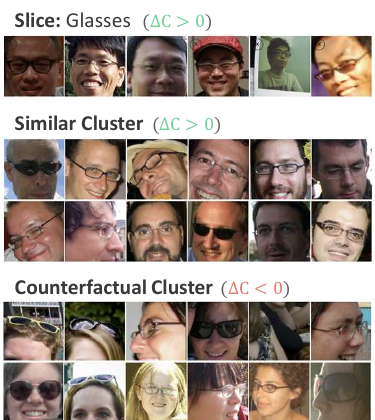}
    \caption{Similar and counterfactual clusters for a slice capturing an unintended subset of \myquote{glasses} for the query $C_b =$ \myquote{A photo of a person}, $C_a =$ \myquote{A photo of a \underline{CEO}.} While the similar clusters display additional masculine presenting glasses-wearers with \dcpos{positive $\Delta C$}, counterfactuals help escape this region by displaying a cluster of feminine presenting glasses-wearers with opposing \dcneg{negative $\Delta C$.}}
    \label{fig:clusters}
    \vspace{-4pt}
\end{figure}

Both similar and counterfactual clusters are updated reactivity as the user gathers samples (and thus change the slice centroid embedding), allowing users to refine and extend their slice in an iterative process. All clusters which contain at least one image already captured by the user are filtered out to avoid displaying samples which have already been considered for that slice. These tools help users find visually similar images to expand their slices and help to guard against users selecting non-representative subsets of the intended visual concept that happen to display bias (as demonstrated in in \figref{fig:clusters}).

\uhdr{Running Example} (\figref{fig:vlslice} \mycirc{C}). From the cluster displaying images of men's suits, the user requests similar clusters and begins adding more images of suits to their slice to expand the slice's coverage of the men's suit visual concept.

\subsection{Validating Model Behavior} \label{sec:validate}

Once users have formed a slice, they can examine the mean $\Delta C$ to draw some conclusions about the model-of-interest's bias. However, it is useful to consider the trend over a larger set of images to determine if it is likely to hold beyond the slice. To achieve this, users can request a correlation scatter plot that charts visual similarity to the slice centroid against $\Delta C$ for each image in the working set. Observing a strong linear relationship in this plot provides additional evidence that the visual concept captured in the slice has a consistent effect on caption affinity. Further, this plot is useful for identifying outlier instances with unexpected behavior (\eg high similarity but different $\Delta C$) which users may add or remove from their slice.

\uhdr{Running Example} (\figref{fig:vlslice} \mycirc{D}). The user requests a correlation plot to validate the bias behavior -- observing a positive slope where visual similarity to the \myquote{men's suit} slice is predictive of higher \myquote{CEO}-ness under the model-of-interest.
\section{\vlslice User Study}
To evaluate \vlslice, we conduct a user study comparing it with a baseline interface, ListSort, representative of typical strategies used to explore model behavior.

\xhdr{Baseline: ListSort.}
The ListSort baseline interface uses the same query inputs and list construction as \vlslice, but several key components are removed. The interface simply sorts images by their change in augmented caption percentile ($\Delta C$), then displays those ranked images to the user without any clustering or human-in-the-loop interactions. The user does not have the ability to view similar or counterfactual images, cannot do additional sorting or filtering, and cannot view the correlation scatter plot. We chose this interface as the baseline because there is no existing tool designed for the task \vlslice supports, and we design the ListSort interface as a representation of the current workflow of ML practitioners. They often simply visualize a sorted list of results to evaluate behavior or use a search engine to manually gather samples for population subsets~\cite{ross_measuring_2020}.

\subsection{Image Data and Model}
\xhdrflat{Data.} 
We use OpenImages \cite{OpenImages} images as our base image set. As our user study focuses on objects and entities rather than scenes, we extract bounding images around annotated objects in the dataset.
We apply non-maximum suppression to the ground-truth boxes to reduce redundant overlapping detections, then extract a square box with side length $1.1 * max(bbox_h,\ bbox_w)$ centered on the detection to capture both the detection and its context. Detections on image borders are padded with the mean RGB value of the dataset. Detections smaller than $64^2$ pixels and those which capture a region bounded by another detection from it's parent in the class hierarchy (\eg, a nose detection on a face) are discarded. In total, this yields a dataset of $I_a = 8.1$ million images. Neither \vlslice or ListSort use the image labels.

\xhdr{Model-of-Interest.} We use CLIP \cite{clip} as a representative vision-and-language alignment model -- using the common variant utilizing a vision transformer \cite{Dosovitskiy2021AnImage} with sequence length sixteen from HuggingFace Transformers \cite{wolf-etal-2020-transformers}.

\subsection{Protocol}
We perform a between-subjects study ($n{=}22$) for \vlslice. Each participant is randomly assigned to use either the \vlslice or ListSort interface and are instructed to complete two tasks using their assigned interface.

\xhdr{Participants.} We recruited 22 participants using departmental mailing lists and word of mouth. Five self-identified as women, fifteen as men, and two as non-binary or other genders. The average age of participants was 27 years old, three were most recently enrolled or completed an undergraduate degree, and 19 a graduate degree. Participation was limited to people who have taken three or more AI/ML courses or have at least two years of professional experience in AI/ML (including graduate studies), and are 18 years old or older. Each session had only one participant and all participants joined remotely via video call. All participants were compensated with a \$15 gift card upon completion.

\xhdr{Protocol.} Each participant takes approximately one-hour to complete the study. First, they are presented with a pre-study questionnaire to collect demographic information, familiarity with vision-and-language tasks in machine learning, and prior experience using other tools for analyzing their models results and behaviors. After completing the pre-study questionnaire, they are introduced to the tasks and given a guided demo of the selected interface with the query $C_b =$ \myquote{A photo of a car},  $C_a =$ \myquote{A photo of a \underline{fast} car}. The study facilitator first demonstrates how to construct slices with the interface for the participant, then they switch roles and the participant demos the same task back to the facilitator. Once the participant is comfortable with the interface, they are given fifteen minutes to complete each of their two tasks and asked to ``think aloud'' while doing so. Finally, the participant completes a post-study questionnaire collecting twelve 7-point Likert-scale ratings and three free-form responses evaluating their experience with the interface.

\xhdr{Tasks.}
Participants are given two tasks: (1) $C_b =$ \myquote{A photo of a person}, $C_a =$ \myquote{A photo of a \underline{CEO}}; and (2) $C_b =$ \myquote{A photo of a house}, $C_a =$ \myquote{A photo of a \underline{nice} house} with $k=3,000$ for both. For each, participants are instructed to discover as many slices as possible while attempting to adhere to the desirable properties given in \secref{sec:intro}. They are informed that these slices should contain visually coherent images with consistent response to the augmented caption.

At the end of each task, we save a snapshot of the slices created by the participant. This snapshot contains the names of each slice, what images were added to it, and all images included in the working set for the query. After each session, recordings of the study are reviewed to transcribe comments made by participants and slices captured by them are manually coded into higher-order categories.
\begin{table}[t]
\setlength{\tabcolsep}{3pt}
\centering 
\renewcommand{\arraystretch}{1.5}
\resizebox{1\columnwidth}{!}{
\centering 
\begin{tabular}{lcccccccc}
\toprule
& \multicolumn{4}{c}{\large ListSort} & \multicolumn{4}{c}{\large \vlslice}    \\
\cmidrule(lr){2-5} \cmidrule(lr){6-9} 
           & \small Slices       & \small \# Img. & \small F1 & \small Missed  & \small Slices         & \small \# Img.          & \small F1 & \small Missed  \\
\midrule
Person/CEO & 4.3         & 107 & .42 & 90 & \textbf{4.6}  & \textbf{141} & \textbf{.59} & \textbf{54}   \\
House/Nice & 3.6         &  87 & .20 & \textbf{41} & \textbf{5.1}  & \textbf{211} & \textbf{.46} & 70   \\
\bottomrule
\end{tabular}}
\caption{Average number of slices identified with average total images cataloged into those slices, F1 slice coherency, and missed images representation metrics, aggregated by participant. \textbf{Bold} is better. Participants assigned \vlslice capture more images with higher coherency in both tasks.}
\label{tab:quantative}
\end{table}

\section{Quantitative Results} \label{sec:quant-results}
\xhdrflat{Size and Number of Slices.}
Using the participant snapshots in each task, we evaluate the number of slices identified and the total number of images categorized into them by participants in \tabref{tab:quantative}. We find \vlslice outperforms ListSort in all cases. We hypothesize causes for the difference between our task results in the discussion (\secref{sec:discussion}). To assess statistical significance, we fit linear mixed effect regression models (\texttt{lmer}) of the form: $y{=}w*\mbox{Interface} + \beta_{\mbox{Task}}$
where $y$ is the measured outcome, $\mbox{Interface}$ is a binary variable indicating \vlslice or ListSort, $w$ estimates the effect strength of the interface, and $\beta_{\mbox{Task}}$ is a per-task intercept modeled as a random effect. Under this model, \vlslice results in 84.58 more images ($p{=}0.017$) and 0.5955 more slices ($p{=}0.295$). This suggests using \vlslice yields statistically significantly \textbf{larger} image sets than the baseline.

\xhdr{Coherency.} To evaluate visual coherency of slices captured by participants, we have annotators perform an outlier detection task. For each participant-collected slice, we subsample eight images and randomly select zero-to-two of those images to replace with outliers. Outliers are randomly sampled from other slices captured by the participant and are constrained to images with visual similarity to the slice centroid within one positive standard deviation of the mean of similarities for all candidate images. This helps prevent trivial outliers and ensures that the participant had considered the outlier images during slice formation. For slices with fewer than eight images, no subsampling is performed and slices with fewer than two images are excluded from the analysis. Annotators are then prompted that each slice contains 0-2 outliers and are asked to select them. We compute F1 over all slices for each annotator. Again using a linear mixed effect regression model of the same form, we find that \vlslice results in an increase in 0.215 F1 score ($p{=}0.006$). This suggests \vlslice leads to statistically significantly \textbf{greater coherence} than the baseline.

\xhdr{Representativeness.} We measure representation by asking annotators to identify images that potentially should have been included in each slice but were missed by participants. For each slice, we measure the similarity between the slice centroid and each image in the working set that was not included in that slice. We then subsample 50 of the 100 most similar images. We display all images captured in the slice, the participant's description of the slice, and the subsampled similar images to annotators and ask that they select all images that should have been included in the slice. Again using a linear mixed effect regression model, we find \vlslice reduces the average number of missed images by $2.1$ ($p{=}0.35$) but this result is not significant at 95\% confidence. As \vlslice slices are both larger and more coherent than the baseline (and thus capture more images relevant to the slice), we suspect the non-significance of this result may reflect a lack of sensitivity in this study.

\xhdr{User Ratings.}
We evaluate participants' average response to Likert-scale post-questionnaire ratings in \tabref{tab:likert}. Participants score \vlslice more favorably that ListSort in 10 of 12 questions with both lower scoring cases being relegated to simplicity of learning and using the interface, an expected result considering the short tutorial period during the study and comparably substantial feature set of \vlslice against ListSort. Nine of ten questions scoring higher on average for \vlslice are measured as statistically significant at $95\%$ confidence under a Mann-Whitney U test.

\section{Qualitative Results} \label{sec:discussion}

Below, we highlight several observations from the user study, providing insights about how people use \vlslice.

\xhdr{\vlslice users discover more abstract slices.}
Mapping slices discovered by participants to higher-level categories reveals trends in the types of relationships typically identified while using each interface. 
Participants assigned to ListSort more frequently discover slices capturing visual concepts which are easy to identify from low-level visual features that require little, and often pre-attentive, visual processing. These slices are frequently based in color (\eg, ``black and white''), structure (\eg, ``truncated subject''), and photographic qualities (\eg, ``blurry''), accounting for $17\%$ ($0\%$ \vlslice) of all concepts identified in the Person/CEO task and $45\%$ ($14\%$ \vlslice) of the House/Nice House task for ListSort. 
In contrast, participants leveraging \vlslice  capture slices relating to historically gendered features, skin tone, and age in $62\%$ ($34\%$ ListSort) of Person/CEO task concepts. For the House/Nice House task, \vlslice participants identify concepts relating to housing density, cultural cues, and features indicative of wealth in $33.9\%$ ($7.5\%$ ListSort) of all concepts.
We note that these choices in slices are emergent and that participants were not instructed to search for any specific or socially relevant visual concepts. We show sample slices created by participants using \vlslice for the Person/CEO task in \figref{fig:ex-person}. Additional examples for both tasks are provided in the appendix.

\begin{figure}[t]
     \centering
     \includegraphics[width=\linewidth]{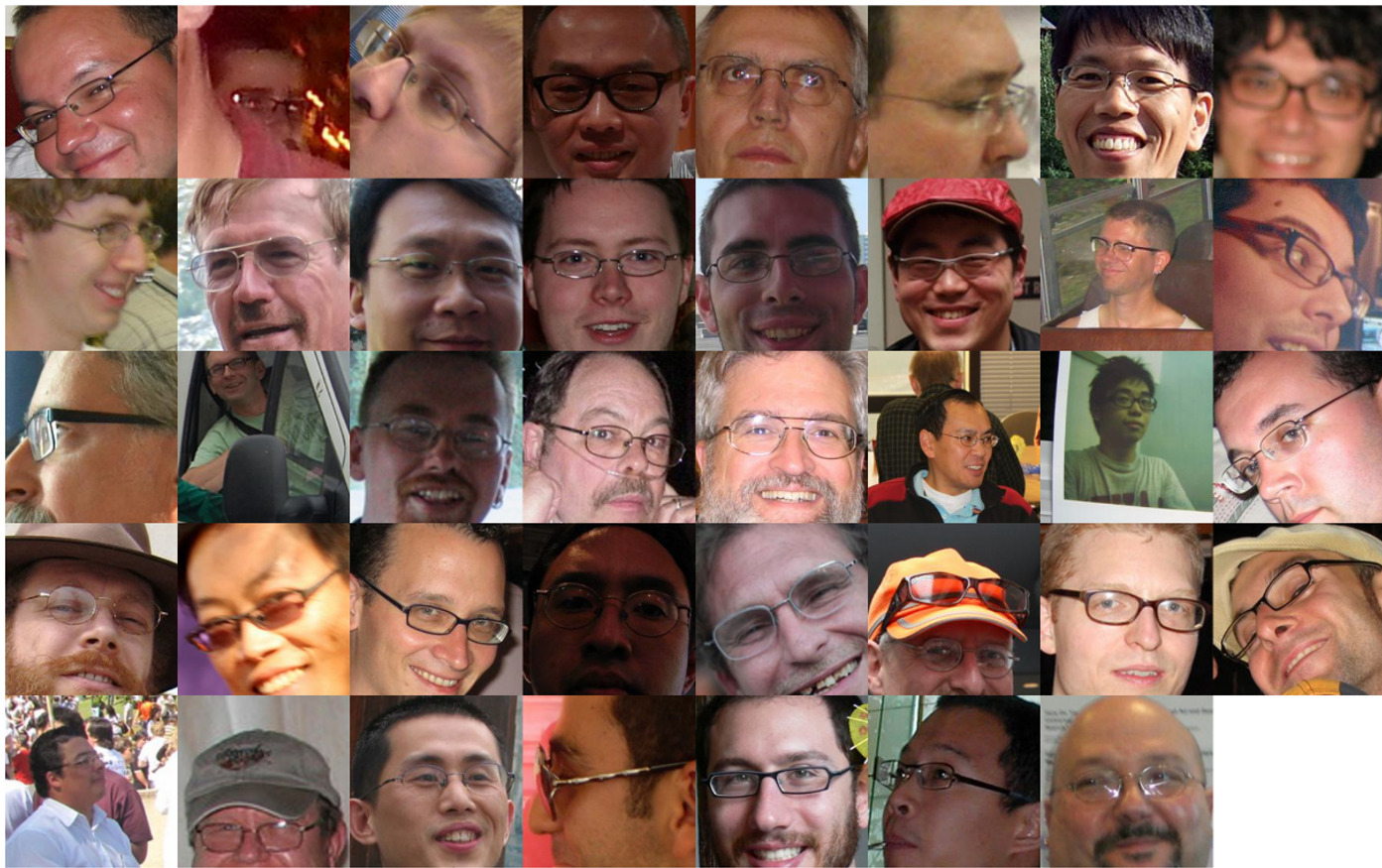}
     ``masculine glasses'' \dcpos{$\Delta C > 0$} \vspace{.4cm}

     \includegraphics[width=\linewidth]{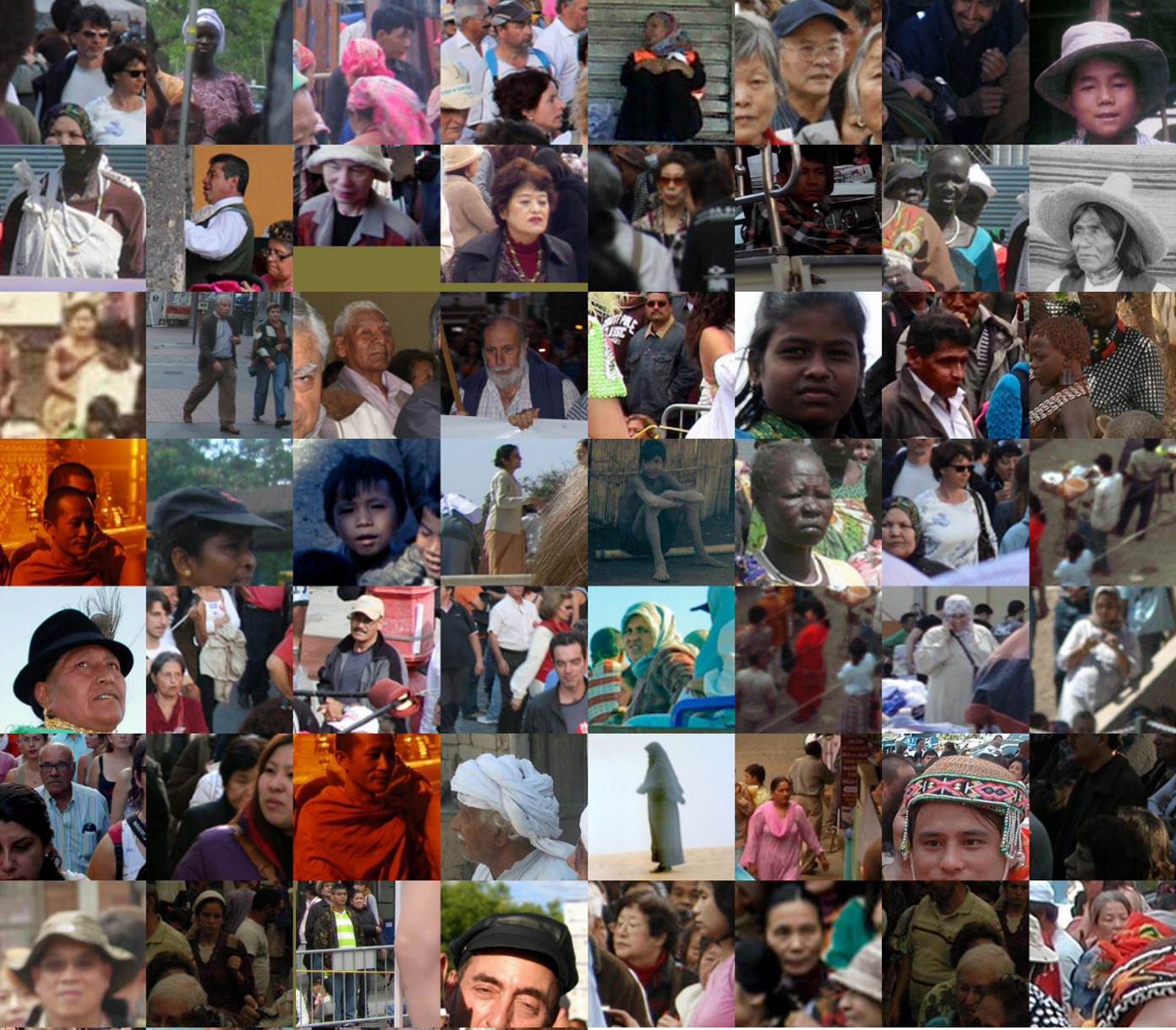}
     ``people of color'' \dcneg{$\Delta C < 0$}
     \caption{Example slices created by participants for the Person/CEO task with \vlslice. In the ``masculine glasses'' slice (top), the participant identified that people wearing glasses with larger features or facial hair have a \dcpos{positive $\Delta C$}, indicating a CEO-like bias. In contrast, the ``people of color'' (bottom) slice has a \dcneg{negative $\Delta C$}, indicating bias against people with darker skin tones being CEO-like.}
     \label{fig:ex-person}
\end{figure}

\begin{table}[t]
\centering
\renewcommand{\arraystretch}{1.25}
\begin{tabular}{@{}lcc@{}}
\toprule
Question                             & ListSort      & \textit{VLSlice}        \\ \midrule
Easy to learn how to use             & \textbf{6.27} & 5.55           \\
Easy to use                          & \textbf{6.00} & 5.55           \\
Confident when using the tool        & 5.45          & \textbf{5.73}  \\
Enjoyed using the tool               & 5.27          & \textbf{6.45*} \\
Would like to use again              & 4.73          & \textbf{6.45*} \\
Image sets capture intended concept  & 4.36          & \textbf{5.36*}  \\
Helpful for finding new behavior     & 5.09          & \textbf{6.55*} \\
Helpful for confirming behavior      & 4.82          & \textbf{6.27*} \\
Easy to build sets of images         & 5.00          & \textbf{6.27*}  \\
Easy to discover additional images   & 4.82          & \textbf{6.36*} \\
My image sets are coherent           & 5.09          & \textbf{5.64*}  \\
My image sets capture systemic bias  & 5.00          & \textbf{6.27*}  \\ \bottomrule
\end{tabular}
\caption{Participants' average 7-point Likert-scale rating for each interface. \vlslice outscores ListSort in 10 of 12 questions. * indicates statistical significance at 95\% confidence in a one-sided Mann-Whitney U test.}
\label{tab:likert}
\end{table}

\xhdr{\vlslice promotes iterative refinement of slices.}
We find that \textbf{all} participants assigned the \vlslice interface engage in iteratively improving the quality of cluster recommendations by utilizing a feedback loop. Participants typically first identify a small number of relevant images from the cluster display and add it to a slice to bootstrap recommendations. Pointing out three neighboring clusters, participant P5 stated \textit{``Some of these look a lot like houses in the neighborhoods I grew up in, lower-income in India specifically''} and selected a small subset of images from each cluster continuing to say \textit{``I can see that it picks up on some of the visual cues individually, but struggles to put them all together''} before adding his selection to a slice and expanding the \emph{similar clusters} display. The most similar clusters provided additional relevant samples and, as P5 continued to update the slice, was satisfactory in support for discovering an otherwise difficult to identify set of images. Likewise, participant P25 stated \textit{``I liked how I could further refine my sets of images. This made it easy to quickly build sets with similar attributes,''} with respect to his cluster recommendations. We anticipate this workflow as the source of the increase in images captured as reported in \tabref{tab:quantative}.

\xhdr{Counterfactuals and correlation plots improve coherency and confidence.}
The above iteration is frequently followed by using counterfactuals to diagnose if the participant's slice has fallen into a systemic positive or negative subset of the visual concept they wish to capture. Participants typically add counterfactual samples until exhausted. For slices with a large support and high variance in $\Delta C$, some participants iteratively switch between similar and counterfactual images until both are exhausted of relevant samples. Participant P1 leverages the correlation plot commenting that \textit{``I'm using the correlation to try and find outliers that I missed when looking through similar and counterfactual photos''} and that \textit{``when I'm looking at this [correlation plot], I'm checking to see if the model really knows the concept I'm trying to capture''} continuing to clarify that they are looking for a steep regression to imply bias and proximal in-concept images most similar to the slices centroid for coherency. We speculate that this workflow helps to improve coherency as reported in \secref{sec:quant-results}.

\xhdr{\vlslice accommodates users regardless of their familiarity with bias dimension.} 
\vlslice can successfully accommodate participants who are both familiar and unfamiliar with the bias dimension.
When a participant is familiar with the dimension, they ask directed questions about behavior with respect to some preconceived notion. For example, participant P6 investigated the intersection of historically gendered features and skin tone in the Person/CEO task stating \myquote{I have a few biases in mind that I'm already aware of and want to search for, so I'm going to [...] to see what the model thinks of them.}
When a participant is unfamiliar with the bias dimension, they seek support from the interface for prompting visual concepts. Several participants find that the initial clustering primes them to investigate different visual concepts they would not have otherwise thought of. P6 stated that \textit{``[filtering and sorting the cluster display] is nice because I'm not really sure what things a house would be biased against, but [VLSlice] primes me to explore some directions.''} This flexibility makes \vlslice an effective tool for both directed bias search and discovery.

\xhdr{ListSort result in unfounded confidence.} 
Although rated lower than \vlslice, participants assigned ListSort are still very confident overall in the coherency of slices they capture (\tabref{tab:likert}). However, our quantitative results (\secref{sec:quant-results}) show that this is not the case. The common methodology of analysis may be leading participants to unsubstantiated conclusions about their model behavior. Specifically, some participants assigned ListSort identify a visual concept with many neighboring samples in the interface, select all those samples for a slice, then continue to a new slice. Examining slices discovered in this case reveals that they often capture a subset of the visual concept targeted by the participant. For example, many participants label their slice as \myquote{formal wear} but captures only images of masculine presenting people wearing suits. Conversely, participant P17 was assigned \vlslice and arrived at a similar result, but after using the provided exploratory tools (\eg, counterfactuals) determined that he was capturing the concept ``men in suits'' instead, proceeding to search for feminine formal apparel as a second distinct slice.

\section{Limitations}
\label{sec:limitations}
\xhdrflat{Choosing an appropriate top-k.}
Users are faced with an implicit sensitivity-specificity trade-off for identifying the domain of subjects with baseline caption $C_b$ through their choice in $k$ for working set filtering. A small $k$ has high specificity for filtering to this domain but poor sensitivity, potentially excluding informative samples. Conversely, a large $k$ has low specificity and high sensitivity, potentially capturing irrelevant samples which will be out-of-distribution for the $\Delta C$ metric. We hope for future work to explore interactive processes at the working set boundary to intelligently select an appropriate value of $k$.

\xhdr{Working-set level model bias.}
While \vlslice allows users to discover biases falling within the working set of samples, it is unable to discover biases along a subject the model is unable to identify or where the bias presented in $C_a$ correlates with a bias against the subjects identified in $C_b$. For example, given a model which is unable to effectively count, and baseline caption $C_b =$ \myquote{A photo of two people}, the model will fail to identify the subject domain. If given the baseline caption $C_b =$ \myquote{A photo of a meal}, augmented caption $C_a =$ \myquote{A photo of a \underline{healthy} meal}, and a model trained from image-caption pairs scrapped from the web, then the notion of both what a meal and healthy meal are will likely be confounded with western cultural norms and thus unable to capture working set examples for evaluating the bias term \myquote{healthy.} We therefore advise that slices identified with \vlslice have high precision and are likely to be representative of a biased notion learned by the model, but potentially poor recall in either case described above and should not be used to argue the absence or non-existence of a bias.

\xhdrflat{Behavior with strongly biased subgroups.}
In the case of a model with strong bias towards some subgroup, but that is still effective for capturing that subgroup in the working set (\eg feminine presentation in the Person/CEO task), there are two ways we may hypothesize \vlslice to change. First, we suspect the \mycirc{B}~\textbf{Explore} step cluster presentation to be more likely to bifurcate along subgroup dimensions, forming additional clusters which capture the subgroup with high magnitude $\Delta C$. Second, during the \mycirc{B}~\textbf{Refine} step, we suspect these bifurcated clusters to be highly ranked within counterfactual cluster recommendations. For example, participants studying people wearing glasses in the Person/CEO task frequently found the subgroup was bifurcated by gender presentation, then discovered the two cluster components using counterfactual recommendations. If model bias is orthogonal to the biases targeted for evaluation, increased user effort may be needed during the refine step to guide the model away from recommendations capturing the disruptive bias instead of the one targeted by the user. For example, by searching through more counterfactuals and bootstrapping \vlslice recommendations for additional steps.

\xhdrflat{Computational complexity for joint encoders.}
As presented, \vlslice is limited to models which compute independent representations of language and imagery, which are used to compute similarity for clustering and $\Delta C$ calculations efficiently. Hypothetically, these
same similarities could be computed as the output of joint encoder models (\eg ViLBERT \cite{vilbert}), but at high computational expense.

\section{Conclusion}
In this work, we proposed \vlslice, an interactive system to discover slices from unlabeled collections of images. We conducted a between-subjects user study to evaluate the effectiveness of \vlslice against common methodologies for identifying model behavior. The results indicate that \vlslice outperforms the baseline for the number of images captured and slice coherency in both tasks. Additionally, participants rate it more favorable than the baseline in 10 of 12 Likert-scale questions describing usability and user confidence in desirable slice properties. We discuss the results of the study and find \vlslice to better support user workflows and promote discovering high quality abstract slices.

{\small
\bibliographystyle{style/ieee_fullname}
\bibliography{main}
}

\clearpage

\begin{figure*}[h!]
    \centering
    \includegraphics[trim=2.2in 1.9in 2.1in 1.9in,clip,width=0.7\textwidth]{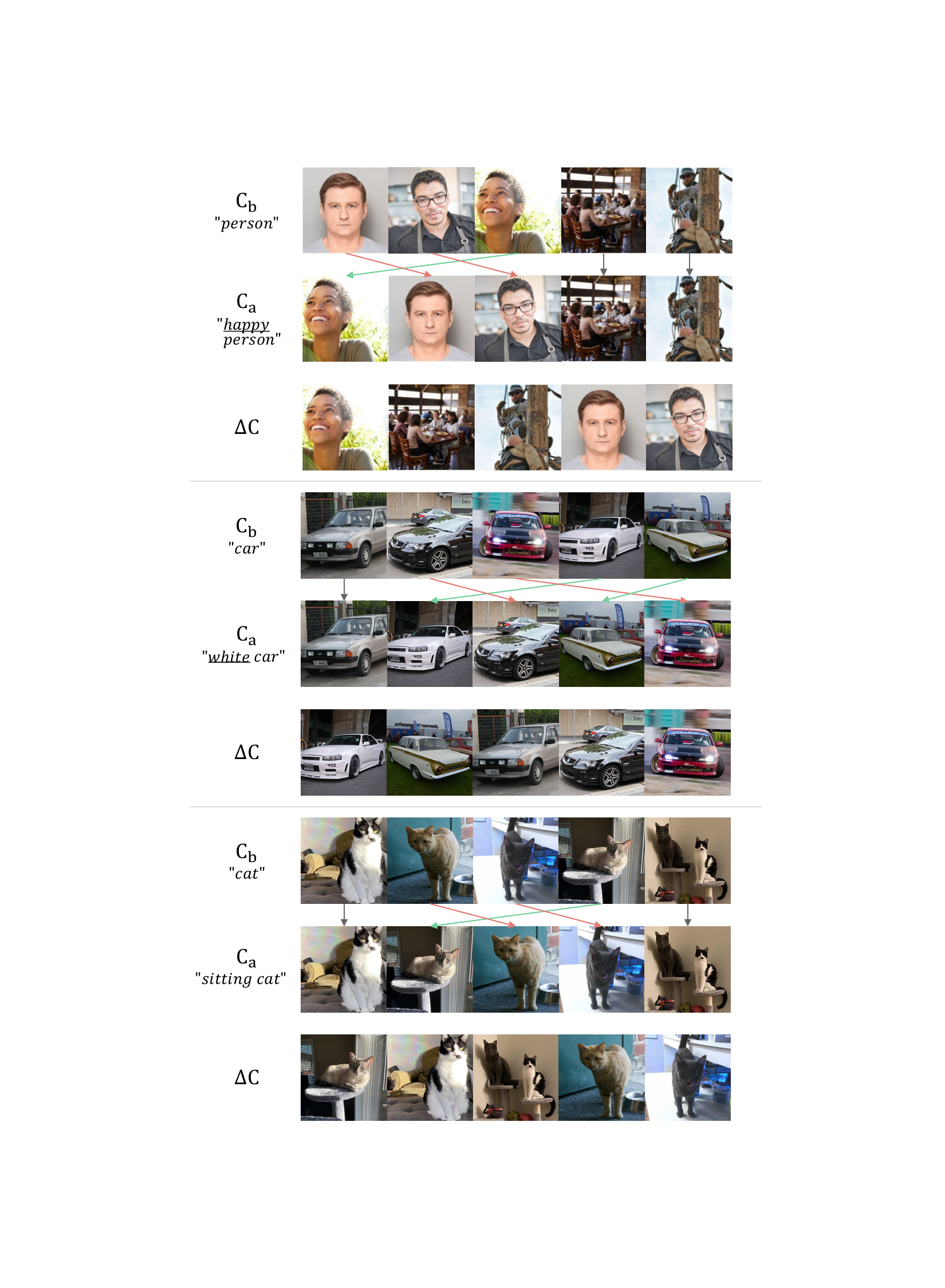}
    \caption{Additional $\Delta C$ ranking examples where $C_b$ and $C_a$ are the baseline and augmented captions, respectively.}
    \label{fig:deltac-ext}
\end{figure*}

\begin{figure}[h!]
     \centering
     \includegraphics[width=0.89\linewidth]{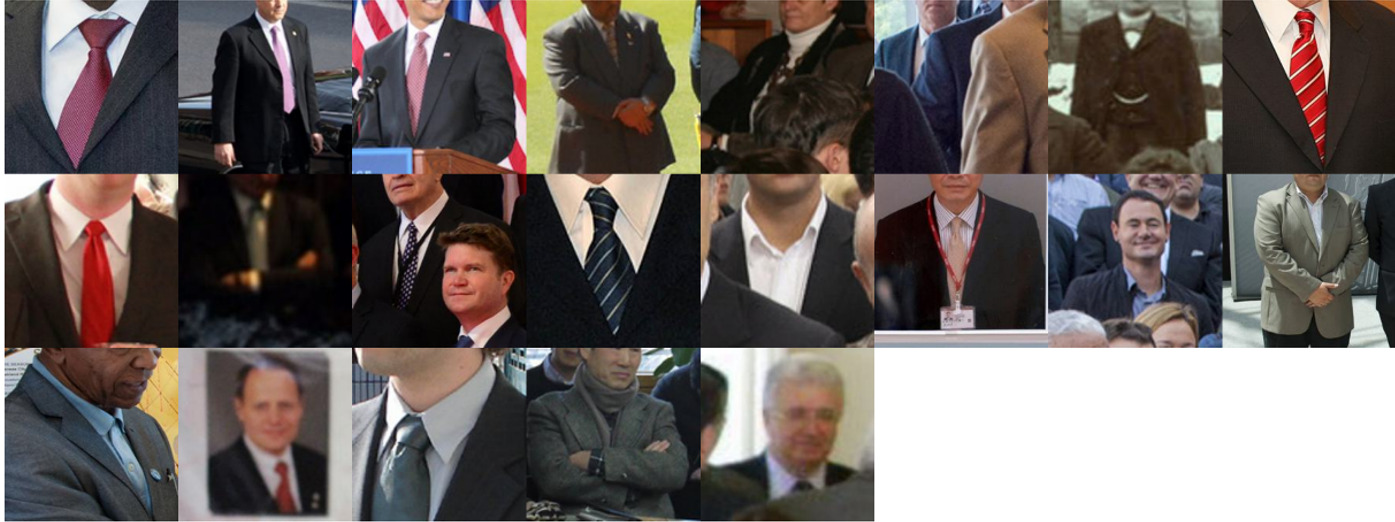}
     ``suits'' \dcpos{$\Delta C > 0$} \vspace{.5cm}

     \includegraphics[width=0.89\linewidth]{images/slices/masculine_galsses.jpg}
     ``masculine glasses'' \dcpos{$\Delta C > 0$} \vspace{.5cm}

     \includegraphics[width=0.89\linewidth]{images/slices/people_of_color.jpg}
     ``people of color'' \dcneg{$\Delta C < 0$} \vspace{.5cm}

     \includegraphics[width=0.89\linewidth]{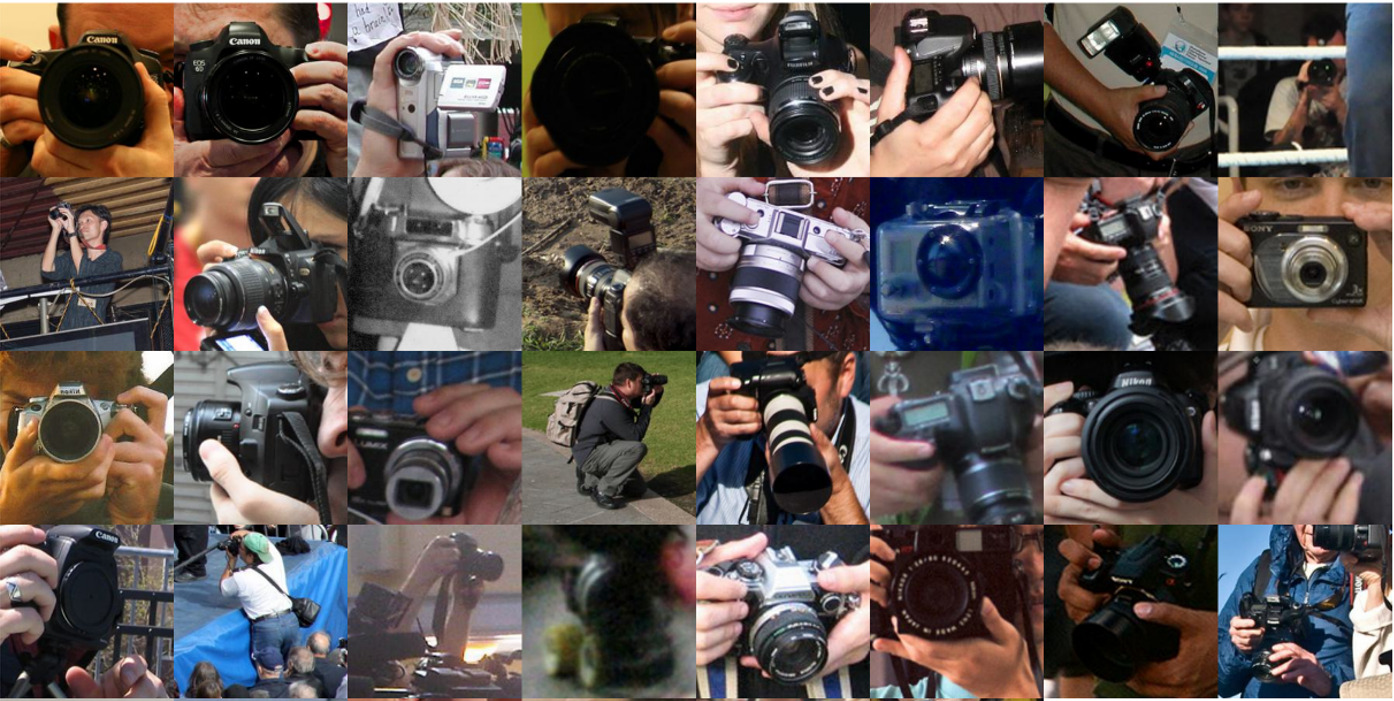}
     ``cameras'' \dcneu{$\Delta C \approx 0$}

     \caption{Additional example slices created by participants for the Person/CEO task with \vlslice.}
     \label{fig:slices-person}
\end{figure}

\begin{figure}[h!]
     \centering
    \includegraphics[width=0.89\linewidth]{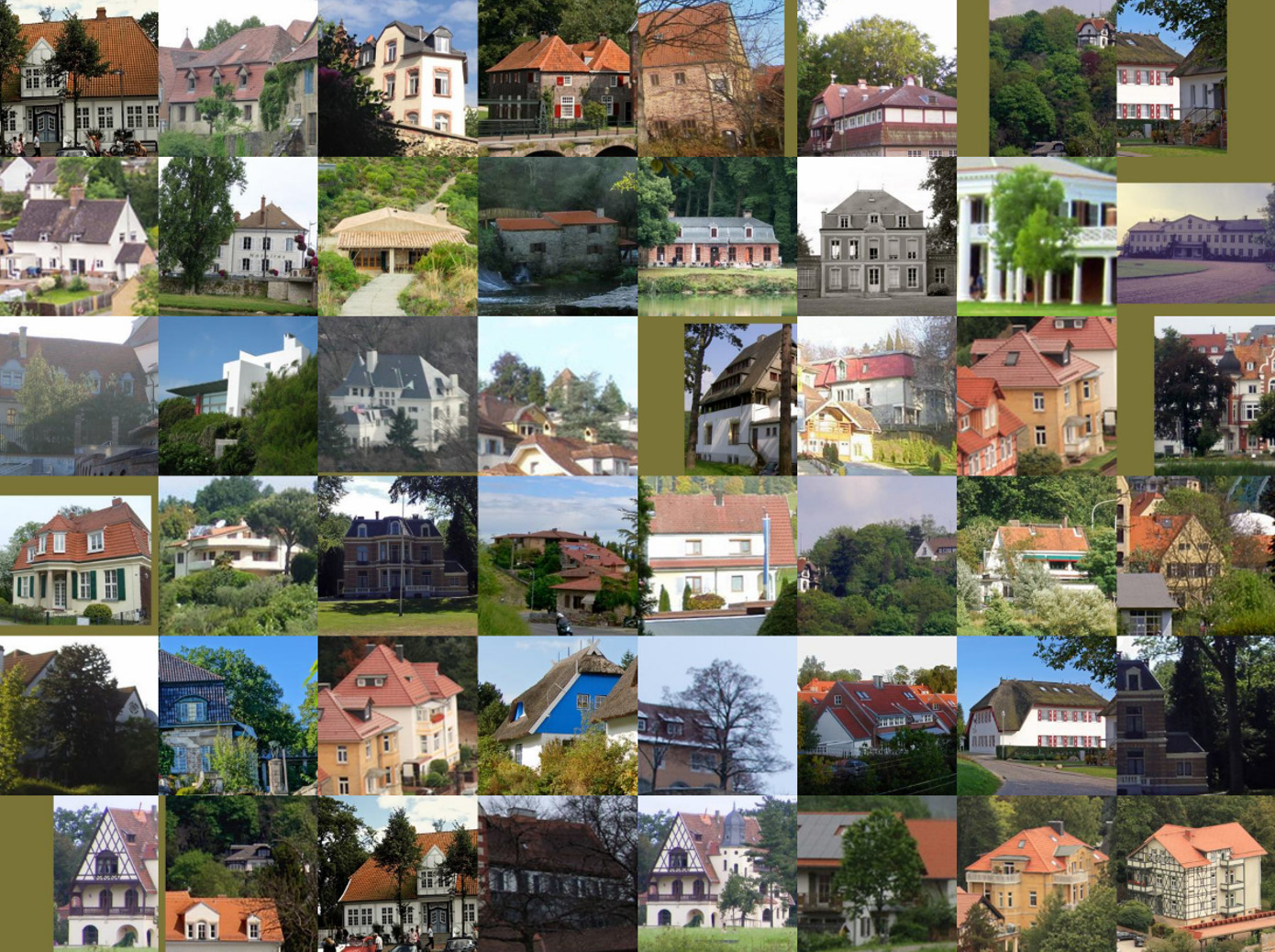}
     ``large european houses''  \dcpos{$\Delta C > 0$} \vspace{.5cm}

     \includegraphics[width=0.89\linewidth]{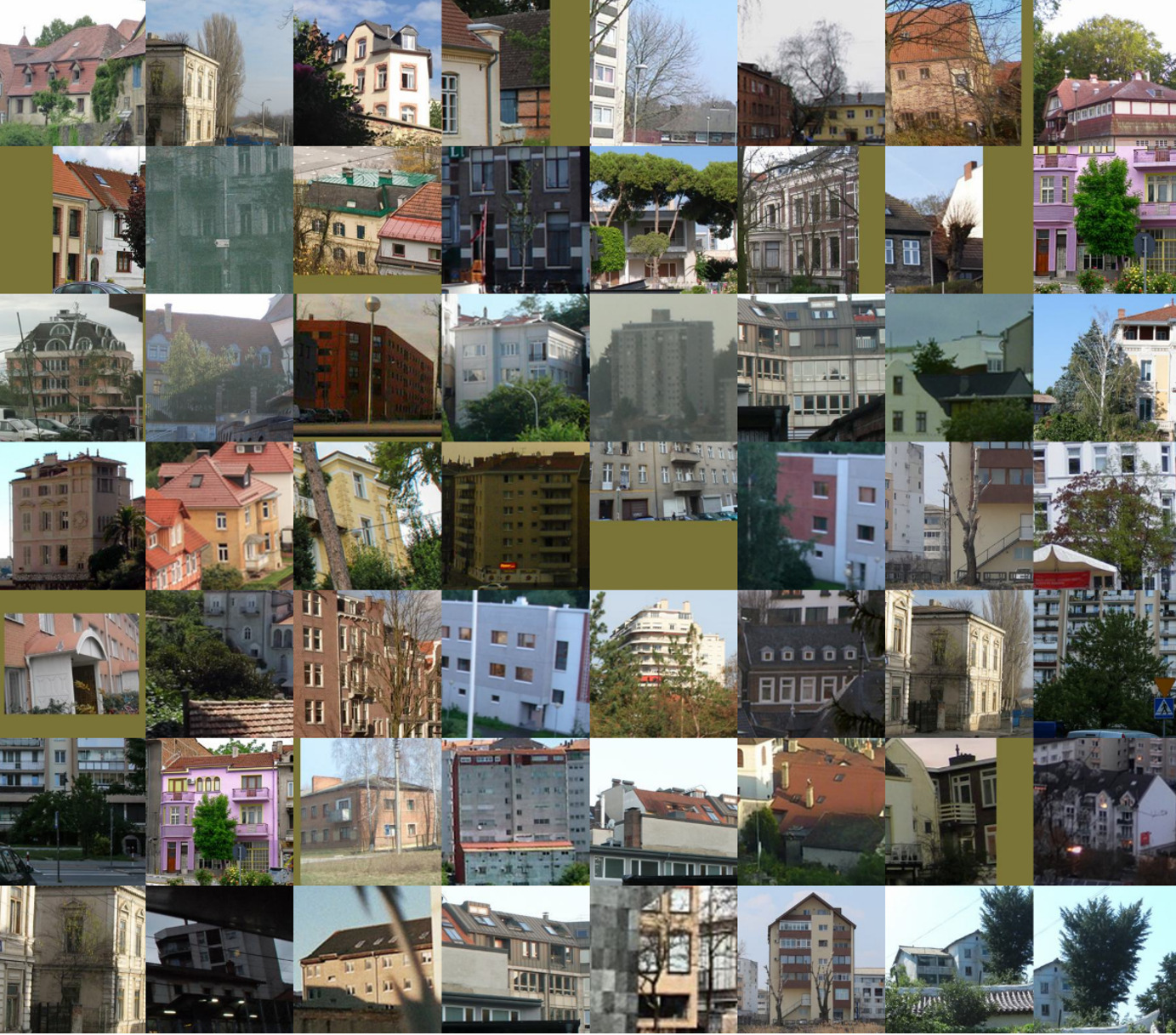}
     ``apartments''  \dcneg{$\Delta C < 0$} \vspace{.5cm}

     \includegraphics[width=0.89\linewidth]{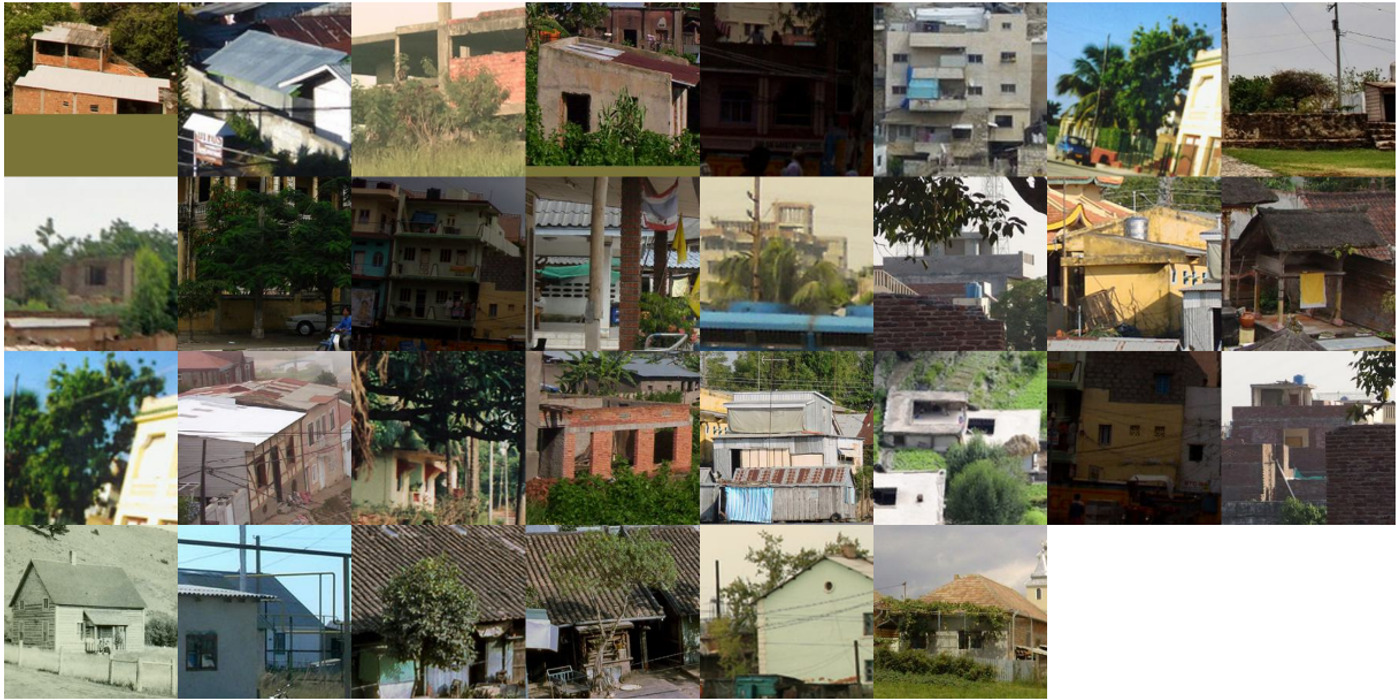}
     ``low-income indian neighborhoods'' \dcneg{$\Delta C < 0$}  \vspace{.5cm}

     \includegraphics[width=0.89\linewidth]{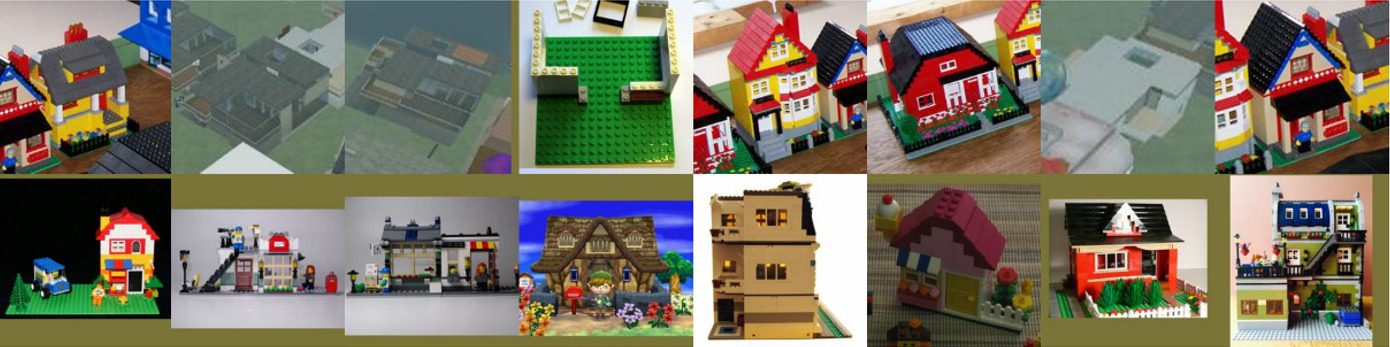}
     ``legos'' \dcneu{$\Delta C \approx 0$}

     \caption{Additional example slices created by participants for the House/Nice House task with \vlslice.}
     \label{fig:slices-house}
\end{figure}

\begin{figure*}[h!]
     \centering
     \includegraphics[width=0.98\linewidth]{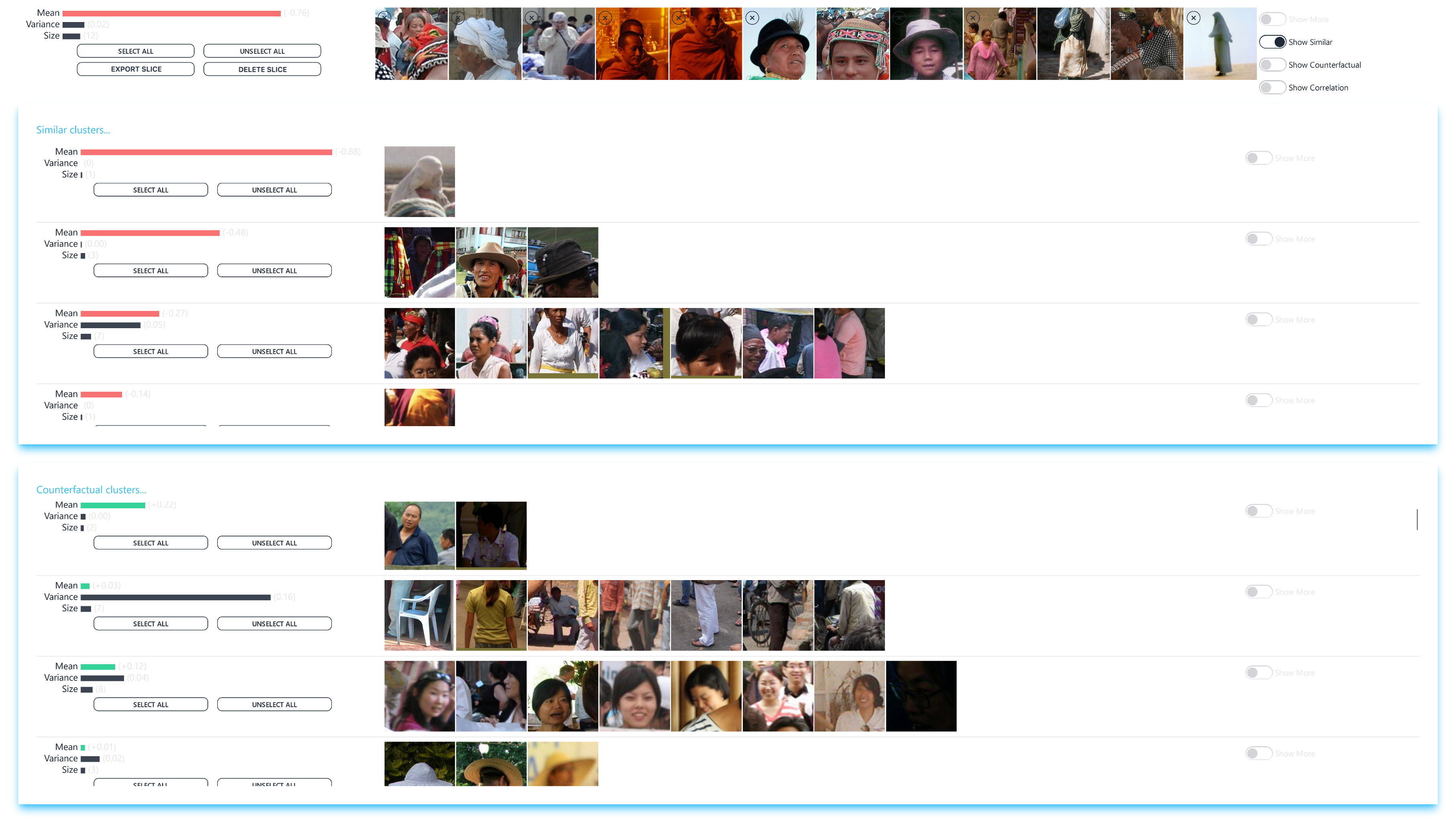}
     ``non-western clothing''  \dcneg{$\Delta C < 0$} \vspace{.5cm}

     \includegraphics[width=0.98\linewidth]{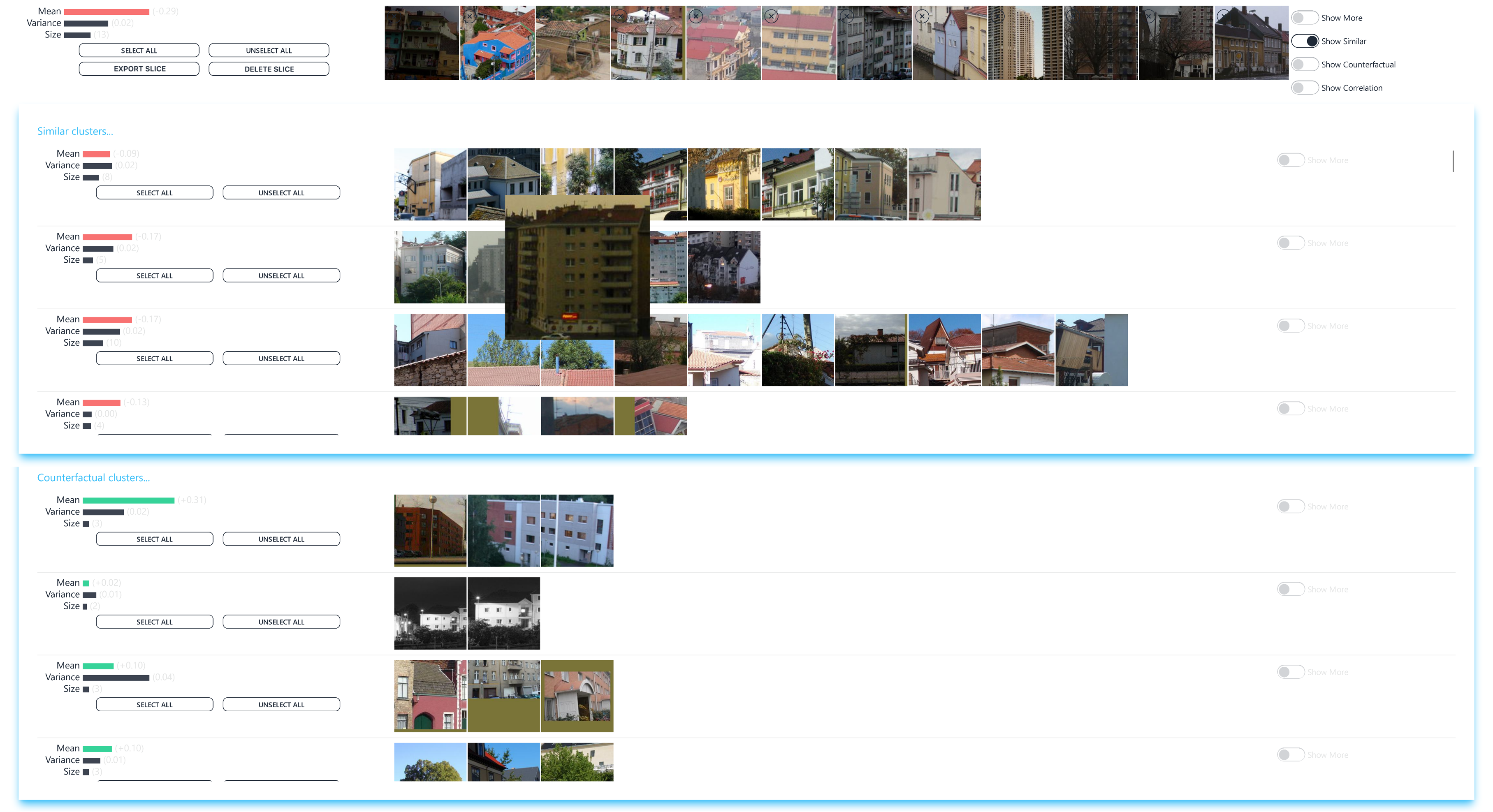}
     ``apartments''  \dcneg{$\Delta C < 0$} \vspace{.5cm}

     \caption{\vlslice similar and counterfactual cluster recommendation interface examples.}
     \label{fig:similar-ext}
\end{figure*}

\begin{table*}[bp!]
\centering
\begin{tabular}{@{}lll@{}}
\toprule
\textbf{Question}                                                & \textbf{Response Type} & \textbf{} \\ 
\midrule
Do you consent to the study?                                     & yes / no               & \\
What is your age?                                                & free response          & \\
What is your gender?                                             & select all that apply  & \textit{(\eg non-binary, woman)} \\
What are your pronouns?                                          & select all that apply  & \textit{(\eg they/them, he/him)} \\
What is your most recent degree program?                         & free response          & \\
Do you have at least two years of professional AI/ML experience? & free response          & \\
Have you taken three or more AI/ML courses?                      & yes / no               & \\
Please list all AI/ML related courses.                           & free response          & \\
What is your expertise level in AI/ML?                           & scale (0 - 5)          & \\
Do you have ViL experience?                                      & select all that apply  & \textit{(\eg ViL navigation, VQA)} \\
Describe your experience with the one(s) above.                  & free response          & \\
Have you used any tools or libraries for analyzing ViL behavior? & yes / no               & \\ 
Which of the following tools/libraries have you used?            & select all that apply  & \textit{(\eg TensorBoard, matplotlib)} \\ 
Can you tell us why you used it and for what purpose?            & free response          & \\
\bottomrule
\end{tabular}
\caption{Pre-study questions and response types given before the interface tutorial.}
\label{tab:prequestion}
\end{table*}

\begin{table*}[bp!]
\centering
\begin{tabular}{@{}lll@{}}
\toprule
\textbf{Question}                                                & \textbf{Response Type} & \textbf{} \\ 
\midrule
The tool was easy to learn how to use.                                                       & Likert (1 - 7) & \\
The tool was easy to use.                                                                    & Likert (1 - 7) & \\
I felt confident when using the tool.                                                        & Likert (1 - 7) & \\
I enjoyed using the tool.                                                                    & Likert (1 - 7) & \\
I would like to use a tool like this one again.                                              & Likert (1 - 7) & \\
I am confident the image sets I created with this tool capture my intent.                    & Likert (1 - 7) & \\
This tool is helpful for finding new model behavior.                                         & Likert (1 - 7) & \\
This tool is helpful for confirming my understanding of model behavior.                      & Likert (1 - 7) & \\
It was easy to build sets of images capturing a concept I was looking for.                   & Likert (1 - 7) & \\
It was easy to find additional relevant images to add to my image sets.                      & Likert (1 - 7) & \\
The images within sets I created are coherent with each other.                               & Likert (1 - 7) & \\
The image sets I created capture a systemic biased relationship between inputs to the model. & Likert (1 - 7) & \\
What was your favorite part of using the tool?                                               & free response  & \\
What was the most frustrating part of using the tool?                                        & free response  & \\
Are there any other comments you have about this tool?                                       & free response  & \\
\bottomrule
\end{tabular}
\caption{Post-study questions and response types given after the participant has completed both tasks.}
\label{tab:postquestion}
\end{table*}

\begin{figure*}[h!]
     \centering
     \includegraphics[width=0.98\linewidth]{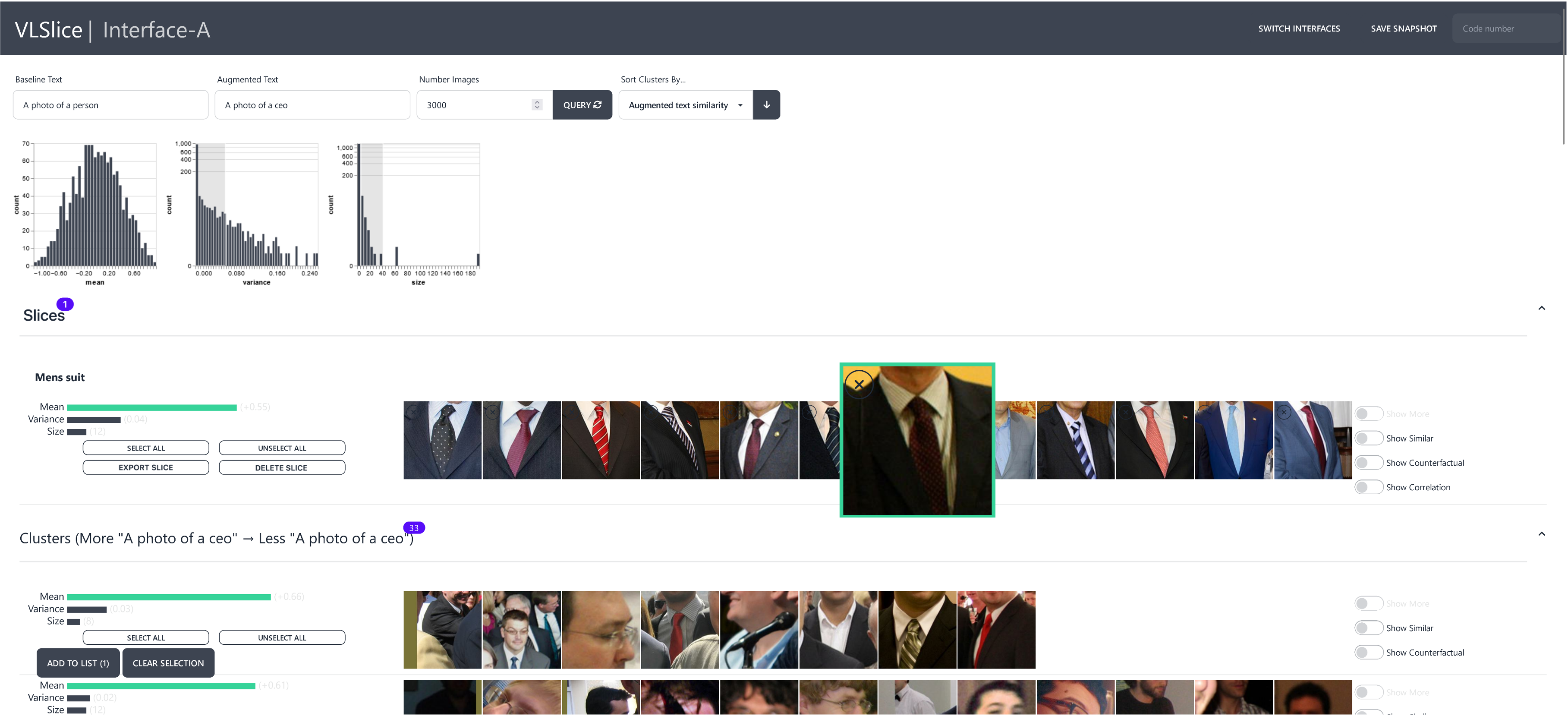}
     \caption{\vlslice interface screenshot. Clicking \myquote{show similar} or \myquote{show counterfactual} expands to display recommendations like those shown in \figref{fig:similar-ext}}
     \label{fig:vlslice-interface}
\end{figure*}

\begin{figure*}[h!]
     \centering
     \includegraphics[width=0.98\linewidth]{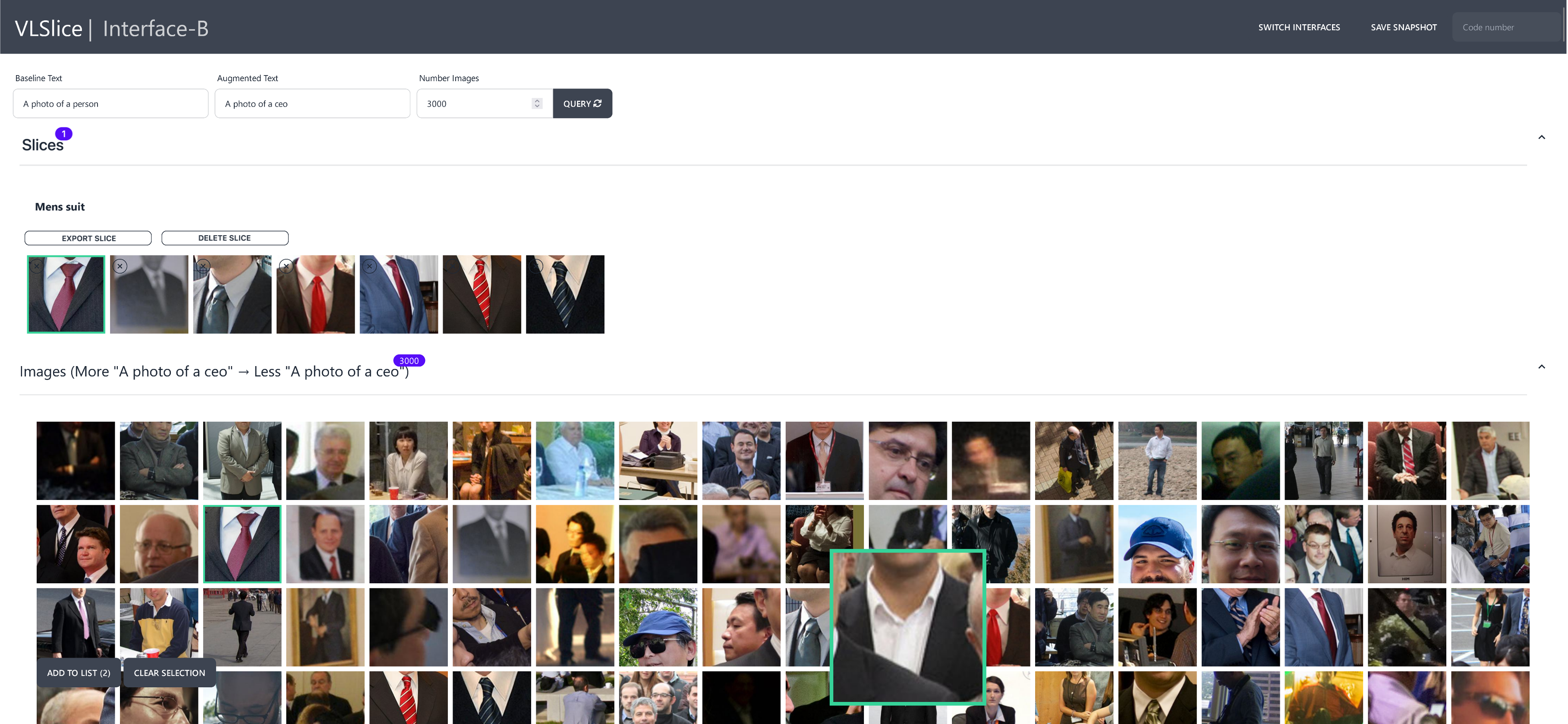}
     \caption{ListSort interface screenshot.}
     \label{fig:listsort-interface}
\end{figure*}

\begin{figure*}[h!]
     \centering
     \includegraphics[trim=0in 0in 0.0in 0.4in,clip,width=0.98\linewidth]{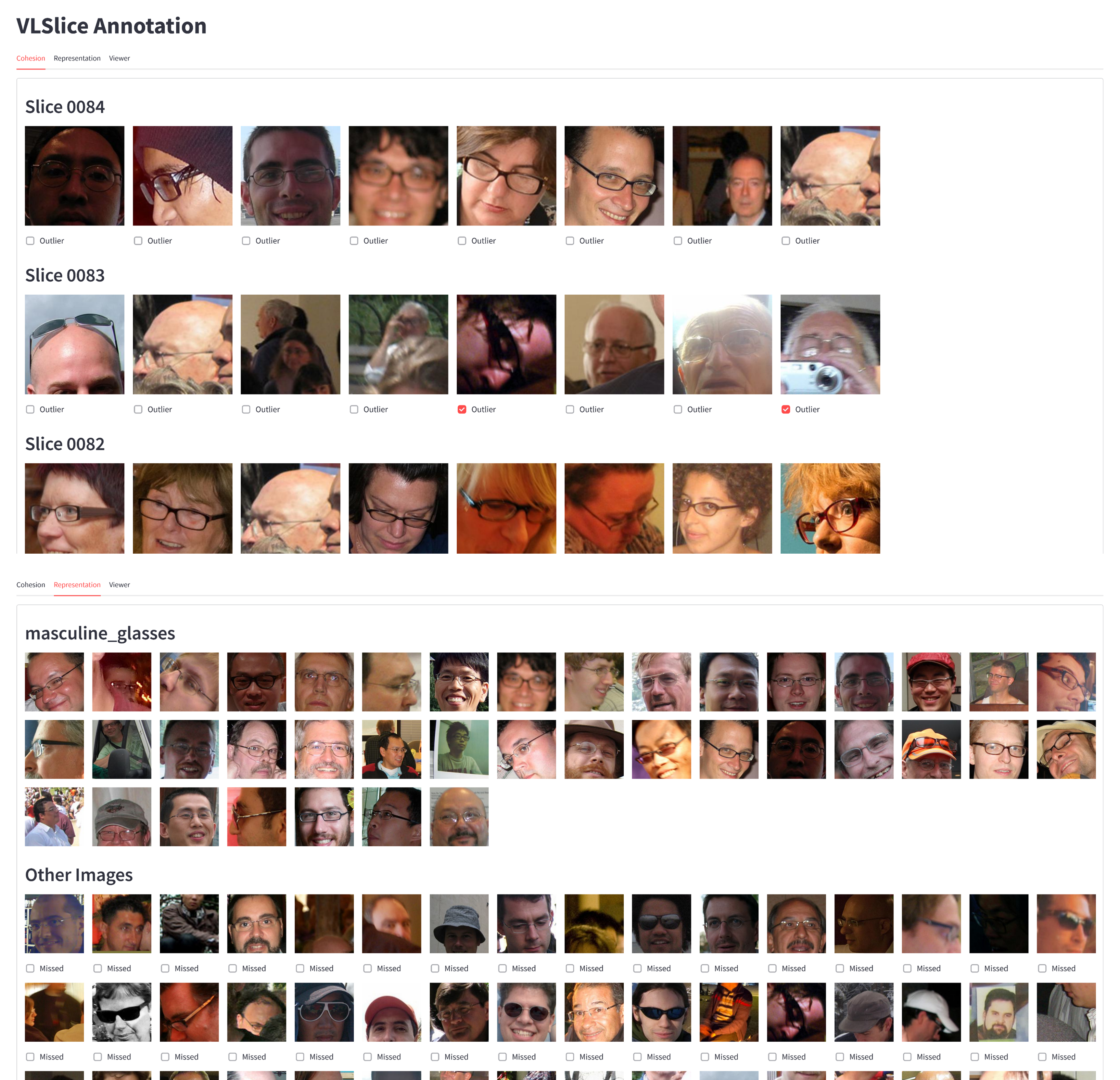}
     \caption{Annotation interface for cohesion (top) and representation (bottom). Annotators select all outlier images for a slice in the first case and any missed images for a slice in the second. No annotator sees the same slice across tasks.}
     \label{fig:annotations}
\end{figure*}

\end{document}